\documentclass[12pt]{IEEEtran}
\usepackage[utf8]{inputenc}
\usepackage{url}
\usepackage{cite}
\usepackage{amsmath,scalerel}
\usepackage{graphicx}
\usepackage{stfloats}
\usepackage{caption}
\usepackage{float}
\usepackage{stfloats}

\title{Real-time Street Human Motion Capture}

\author{
\IEEEauthorblockN{Yanquan Chen\IEEEauthorrefmark{1}, Fei Yang\IEEEauthorrefmark{1}, Tianyu Lang\IEEEauthorrefmark{1}, Guanfang Dong\IEEEauthorrefmark{2}, Anup Basu\IEEEauthorrefmark{3}}\\
\IEEEauthorblockA{\IEEEauthorrefmark{1}University of Alberta. \emph{tlang@ualberta.ca}}
\IEEEauthorblockA{\IEEEauthorrefmark{1}University of Alberta. \emph{yanquan@ualberta.ca}}
\IEEEauthorblockA{\IEEEauthorrefmark{1}University of Alberta. \emph{fei5@ualberta.ca}}
\IEEEauthorblockA{\IEEEauthorrefmark{2}University of Alberta. \emph{guanfang@ualberta.ca}}
\IEEEauthorblockA{\IEEEauthorrefmark{3}University of Alberta. \emph{basu@ualberta.ca}}

}

\date{Dec 2021}

\begin{document}

\maketitle

\begin{abstract}

In recent years, motion capture technology using computers has developed rapidly. Because of its high efficiency and excellent performance, it replaces many traditional methods and is being widely used in many fields. Our project is about street scene video human motion capturing and analysis. The primary goal of the project is to capture the human motion in a video and use the motion information for 3D animation (human) in real-time. We applied a neural network for motion capture and implement it in the unity under a street view scene. By analyzing the motion data, we will have a better estimation of the street condition, which is useful for other high-tech applications such as the self-driving cars.
\end{abstract}

\section{Introduction}
Motion capture is the process to record a series of movements of an object or human being. This technology can be used in a variety of fields, such as military, entertainment, sports, computer vision, and medical applications etc.  If we compare motion capture with some traditional algorithms, the motion capture method is low-latency and can almost achieve real time capture. Also, we can obtain the motion file after capturing. And this motion file can be reused for many purposes. 

There are also many applications related to motion capture. For example, motion capture can be used in the gaming industry. It is very common to capture human motions and then combine these motion files with the characters in games, so the characters in games can actually move like humans. We can also apply motion capture to sports. We can analyze the physical condition, match performance, and also the technical problems of the athletes by studying their motion during practice or games. We may also utilize these motions to prevent injuries. In medical areas, especially physiotherapy, we can capture the patient’s movement during rehabilitation training and then analyze these motions to estimate whether the patient is getting better. Also, based on these motions, to establish a more suitable recovery plan for the patient.

\section{Brief Summary of Existing Work}

\subsubsection{Object detection}
In the motion detection, the first part is to identify the objects and segment the background and objects, so we introduce some segmentation algorithms first.

In 2014, Ross Girshick and his team proposed the RCNN system to detect objects in an image and enclose them with bounding boxes. They used the selective search algorithm to choose around 2000 region proposals, and pass each of these region proposals to a CNN network to create feature vectors. Next, each of these vectors will be passed to a SVM classifier to classify the object. This is a robust algorithm which achieves state of the art results, but is a bit time consuming. \cite{girshick2014rich} One year after the proposal of rcnn, Fast-rcnn was introduced and significantly improved the efficiency of Rcnn. Instead of passing each region proposal to a CNN, the fast-rcnn passes the whole image with region proposals into the CNN and obtains a feature map. After, each region proposal will be passed to a RoI pooling layer, which returns the feature map for each region. The region feature map will then be flattened and passed to fully connected layers to do classification and bounding box predictions. This algorithm significantly improves the training time cost since it only goes through the CNN only for once.\cite{girshick2015fast} Furthermore, the faster-rcnn further improves the performance of fast-rcnn by using the RPN to select region proposals. Instead of selective search, faster-rcnn uses a sliding window over the feature map and generates 9 anchors for each pixel during the scanning process. Then, it applies the RPN to classify the anchors as foreground or background, and also do regression on the bounding boxes prediction. After that, combining the foreground anchors and features and passing them to the RoI pooling layer and Fully connected layers to do classifications.\cite{ren2016faster}
In 2018, kaiming He and his team further improved the performance of faster rcnn by introducing the mask rcnn. Based on the faster rcnn, mask rcnn adds a branch to generate the mask for the region of interest, and changes the loss function to combine classification loss, bounding box regression loss, and the loss of mask. Besides, it also uses a new RoI pooling method called the RoIAlign to uniform the sizes of target cells and apply interpolation to calculate feature values, which significantly improves the overall accuracy.\cite{he2018mask}

Even though deep learning methods for object detection have been rapidly evolving in recent years, there are still some worth mentioning traditional methods that produce decent results. 

Scale-Invariant Feature Transform(SIFT) algorithm which computes the scale-invariant corner features,  it computes the local maxima based on Difference of Gaussian(DoG) for difference scales and space to return a set of potential keypoints with varies scales, which achieve invariance to image scaling. Also, its computation and assignment of orientations for each keypoint achieve the invariance to image rotation.\cite{sift}
The HoG algorithm also classifies the object based on the pixel gradient. It uses an 8x8 grid block and applies it to each section of the image. Next,  it computes the image intensity difference in magnitude and direction within each grid, and stores the information into a histogram which represents the weight of gradient change in each direction. Then, we can show the gradient information for each grid block on the corresponding location of the image, and pass the ‘gradient image’ to a classifier to perform object detection. This method always returns decent results especially for human detection tasks.\cite{hogDalal}
Besides these well-known methods, Tao Xu and his team proposed a modified Gradient Vector Flow Based Active Shape Model(GVF-ASM) to improve the performance of lung area detection and segmentation. They added a “sgn” function to maintain the direction of GVF vectors and an exponential term to gather points to distinctive edges. As a result, the proposed method makes the point evolution process less sensitive to different parameters, and also improves the segmentation performance.\cite{lungXutao} Back to our project, similar to motion detection, Xianting Ke and his/her team proposed the race classification based iris image segmentation method, which included feature detection, classification and segmentation steps. First, it applies LGBP to obtain important features of the image. Then, the extracted iris features are passed to the support vector machine to do classification. If it is a human iris, the method utilizes a circular Hough Transform to localize the interior boundary and an active contour model to localize the exterior, and vice versa for non-human eyes.\cite{IrisKe} Sometimes it is hard to distinguish between real motion and fake motion. Yuan Chengshan and his team somehow gave us some hints by his proposed method on real/fake fingerprints detection. Different from the normal feature extraction method, they utilized three feature extraction approaches which are SIFT, LBP, and HOG, and then applied a feature fusion operation to combine all these extracted features from three different methods. Next, pass the fused features to the SVM to perform classification. The result shows that after the feature fusion operation, the prediction accuracy is generally higher.\cite{fingerprintYuan}
Zhang et al\cite{zhang2010robust}'s research in subtraction showed on possible route for extracting high-quality dynamic foreground layers with relaxed background scene restrictions. A comprehensive system for accurately computing object motion, and depth information are designed. This novel algorithm combines different clues to extract the foreground layer, where a voting-like scheme robust to outliers employed for optimization. The system can handle difficult examples in which the background is non-planar or the camera freely moves during capturing. Here is the pipeline of their algorithms. Firstly, they use GrabCut, a third-party layer extraction tool to extract the foreground from the background. At this point, the rough mask is obtained. Then, they use SIFT to track POI across the frames to determine which region to refine. Lastly, the depth map is recovered using the multiview stereo method. These factors are used to refine the final mask. Despite its high accuracy, some problems remain to solve. The speed is not fast enough. 

Xiong et al.\cite{articleXiong} Purposed a method to automatically extract object's outline from datasets that only contain bounding boxes. In their research,This end-to-end learning framework is for segmenting generic objects in both images and objects. Their approach produces a pixel-level mask for regions that like objects even for object categories never seen during training. They called this feature objectiveness. Beyond the core model, 
another contribution of 
their approach is how it leverages varying strengths of training annotations. They propose ways to exploit labeled data like images that are labeled with rectangles. However, the speed is not indicated in their published paper and the results show that its accuracy does not 
good enough.

Wu et al.\cite{wu2015moving} proposed a computationally efficient algorithm that can detect objects in a given scene. CFG region is extracted by performing reduced SVD on multiple matrices that are from particle trajectories. Then, the BG of pixels is rebuilt by a fast inpainting method. After subtracting the BG, the foreground is extracted by an adaptive thresholding method. Finally, The mean-shift segmentation method is used to refine the foreground. There are three aspects to demonstrate their contribution:1. To determine the moving objects (foreground) in images, a computationally efficient coarse-to-fine thresholding framework is proposed, which neither recovers explicit 3D BG scene nor requires complicated geometric constraints and a probabilistic graphical model for BG/foreground. 2. The proposed block RSVD-based thresholding on particle trajectories allows us to extract the coarse foreground (CFG) efficiently in a general 3D scene. 3. An adaptive thresholding method is presented to handle the misdetection problem in images that contain multiple moving objects in a large range of motion. The advantage of this algorithm is its efficiency and simplicity. But this algorithm requires a strong hypothesis: the objects significantly differently from its surrounding BG.

Some other detection methods listed here are interesting as well. Fiala \cite{FIALA20021863} proposed an object detection by hough transform. It is one of the basic methods to recognize geometric shapes from images in image processing.  The basic principle of Hough transform is to transform a given curve in the original image space into a point in the parameter space by using the duality of point and line.  Thus the problem of detecting a given curve in the original image is transformed into a problem of finding the peak value in the parameter space.  That is, the detection of the overall characteristics into the detection of local characteristics.  Such as straight lines, ellipses, circles, arcs and so on. Cheng \cite{Cheng2007AirwaySA} creatively combined  Gradient Vector Flow Snake  and edge detection methods for image segmentation.Snake is a dynamic contour extraction algorithm. Snake has two forces, external and internal. The internal force is the initial contour itself, and the external force is from the image itself.  
Internal forces have two controls: A, controls tension. B, controls rigidity. Basically,  A controls continuity and smoothness. B guides curve guide boundary. 

Limiting detection regions and using a shape-adaptive model can also improve object detection results. Yin et al.\cite{yin2001nose} proposed a new low-bit rate model-based coding detection method used for human nose detection. This method is verified practicable. Kanade–Lucas–Tomasi feature tracker (KLT) is usually used to track extracted features in a few continuous frames in a video. Singh et al.\cite{singh2005gaussian} introduced using weighting functions to improve KLT. Their experiment verified the gaussian weighting function and the LoG weighting function can improve the performance of KLT. It is a good idea for our motion detection training. YOLO is the abbreviation of "You Only Look Once"\cite{redmon2016you}. Although it is not the most accurate algorithm, it is a good compromise between accuracy and speed.YOLO-v3 implements a Full Convolution Network(FCN) named Darknet-53, which has 53 convolution layers. Each convolution layer is followed by batch normalization layers and leaky ReLU layers.  The YOLO-v3 algorithm uses a single neural network to act on the image, divide the image into multiple regions and predict the bounding box and the probability of each region. YOLO-v3 draws on YOLO-v1 and YOLO-v2. Although there are not many innovations, it improves the detection accuracy while maintaining the speed advantages of the YOLO family, especially for small objects. Liu et al.\cite{liu2016ssd} proposed Single Shot Multibox Detector in 2016. "Single Shot" means it is one kind of one-stage object detection, and "Multibox Detector" means it can detect multiple objects. The main idea of one-stage object detection is to perform dense sampling evenly at different positions of the image. Different scales and aspect ratios can be used when sampling, and then CNN is used to extract features and directly perform classification and regression. This kind of method is really fast but hard to train. Compared with YOLO, SSD overcome YOLO's disadvantage that hard to detect small objects. SSD directly uses CNN to detect instead of detecting after the Fully Connected Layer like YOLO. The advantage of SSD is that 1) it can apply detection at multiple scales data; 2) Faster speed than YOLO; 3) More accuracy than Faster RCNN. 

\subsubsection{Image Segmentation}

Image Segmentation is also one kind of object detection. 

Based on the stability of the texture information, Faraji et al\cite{faraji2018segmentation}. upgrade a region detector called Extremal Regions of Extremum Level(EREL) and propose a region selection strategy to segment lumen and medium in Intravascular Ultrasound(IVUS) image by clustering the regions of interest. EREL is a region detector that detects a series of connected pixels from the image by conjunctly employing a Union-Find structure with the edge information. The authors first use a non-linear median filter to preprocess the dataset in order to decrease the sensibility of the proposed to speckle noise. In our project, we can also consider adding denoise preprocessing to our dataset to avoid the noise influencing the detection. Moreover, adding edge features this paper mentioned extracted by edge detection might improve object detection. They then investigated whether the extracted EREL is considered to be the appropriate region of the lumen and media in the human coronary artery. Next, an ERELs selection strategy was introduced. This strategy is based on creating a vector that represents several textural features of the regions and searching for the higher-textural-stability-score regions. Credit to the previous denoising, this method works very well. The average Hausdorff distance(HD) for both lumen and media is less than 0.3 mm, which is better than other methods. 

Another method for the segmentation of the IVUS image is proposed by Yang et al. The authors proposed a Dual Path U-Net (DPU-Net)\cite{yang2019robust} based on Fully Convolutional Network (FCN) to do automatically segmentation in the IVUS. They use a deep architecture and a dataset augmentation controller to overcome the limitation of small training datasets. Lacking training data is the most common challenge in medical image analysis. In our project, we will use a recent dataset called TACO\cite{proencca2020taco} which only has less than five thousand images. The reasonable use of a small dataset in this paper is worth learning. The deep architecture is built based on the UNet architecture and combined with an improved extension of the IVUSNet\cite{yang2018ivus}. The introduced deep architecture not only has a better generation by training a small group of data but also doesn't require continued training using a pre-trained model. Data augmentation is usually used to solve missing data or lacking data problems. The common method is doing transform on the existing datasets, such as rotation, rescaling, or panning. However, not all argumentation method is useful in medical applications. Redundant augmentation may increase calculation power cost. Thus, the authors applied the combination of traditional transforming methods and three filters they introduced to expand the dataset and improve the accuracy of the final segmentation. These three filters mimic three different intravascular Ultrasound artifacts, bifurcation mask, side vessel mask, and shadow mask. Low computational cost and ability to apply multiple operations on one image make the designed filters better doing data augmentation. 

The DPU-Net they proposed has better performance in the Intersection over Union (IoU) and HD than the two state-of-the-art methods, SegNet\cite{badrinarayanan2017segnet} and UNet\cite{ronneberger2015u}. The DPU-Net has a higher 8–15\% in terms of HD distance than existing methods, and DPU-Net also shows a strong generalization property for predicting.

\section{Methodology}
\begin{figure}[H]
    \centering
    \includegraphics[scale=0.5]{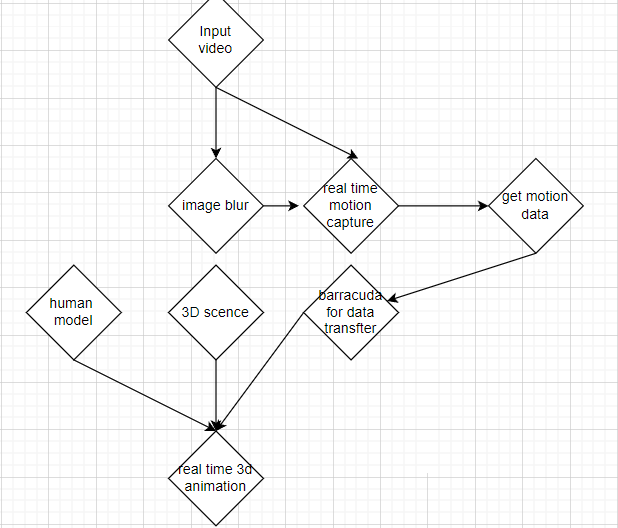}
    \caption{flowchart}
    \label{fig:PP7}
\end{figure}

Basically, as shown below Fig.\ref{fig:PP7}, our project has three stages. 1) Data process 2) Motion capture 3) Model movement.

\subsection{Data process}
\begin{figure}[H]
    \centering
    \includegraphics[scale=0.5]{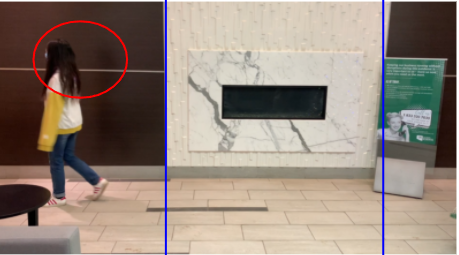}
    \caption{Challenging}
    \label{fig:PPn}
\end{figure}

During the experiment, we get poor motion capture performance if there is a colour overlap between the person and environment. In the below Fig.\ref{fig:PPn}, the colour of hair and wall are similar (red colour). Only blue area performance is good. So we decided to implement video blurring to reduce the influence of the background environment.

\begin{figure}[H]
    \centering
    \includegraphics[scale=0.5]{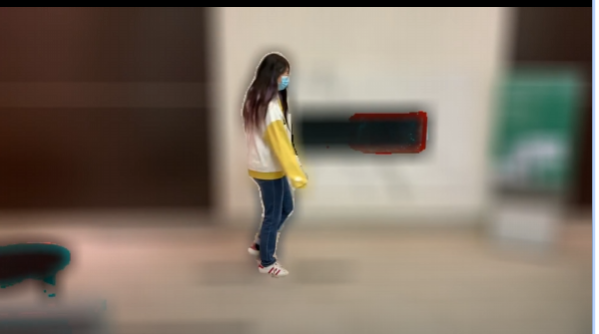}
    \caption{blurring}
    \label{fig:PP2}
\end{figure}

To implement this part we use Xception convolutional neural network which is 71 layers deep and pre-trained model from Matlab. We show one blurring frame from the video below (Fg.\ref{fig:PP2}).

\subsection{Motion capture }
\begin{figure*}[]
    \centering
    \includegraphics[scale=0.2]{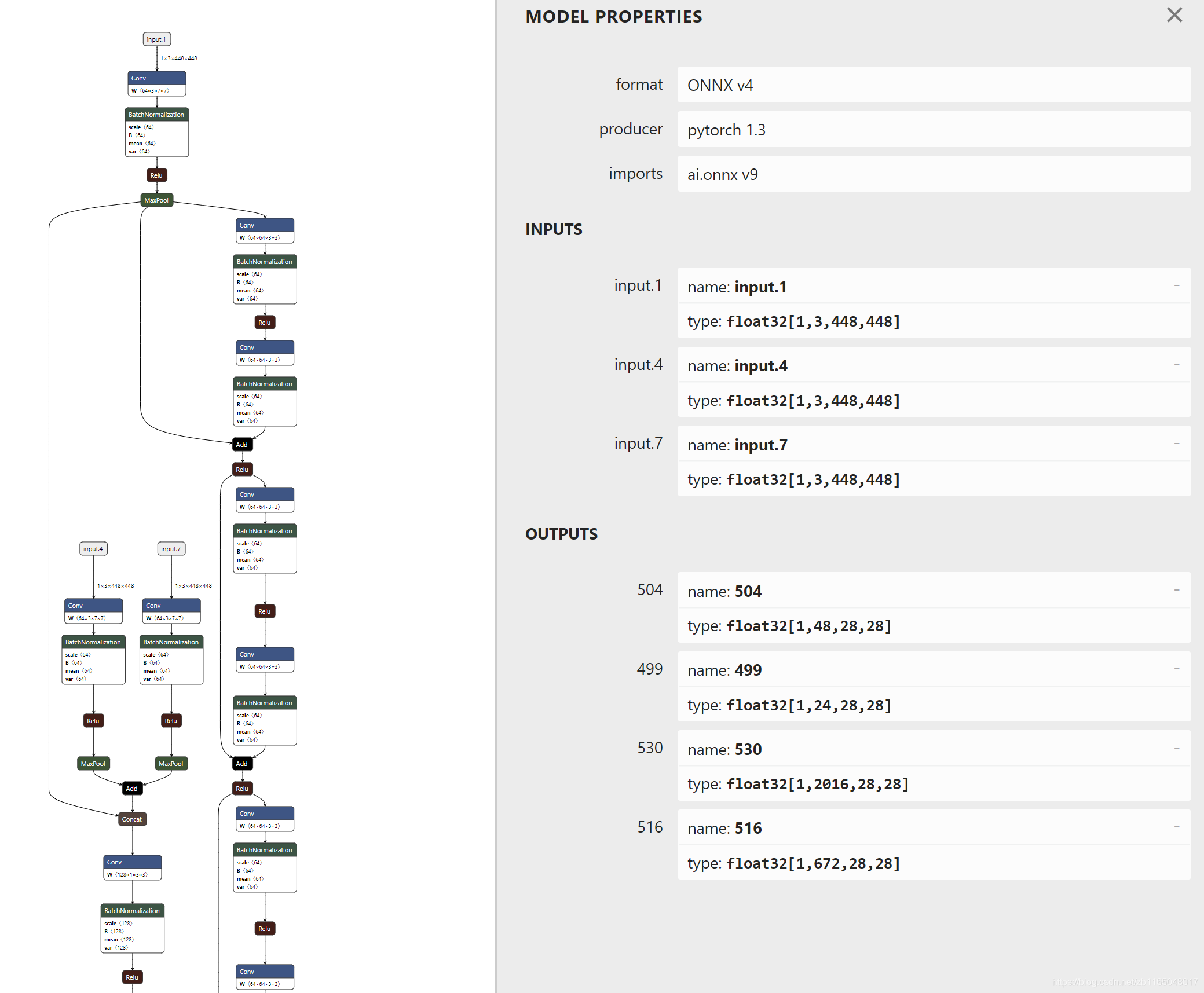}
    \caption{resnet-34}
    \label{fig:PP1}
\end{figure*}

We use ResNet-34 to capture the motion. The input size is (448, 448, 3) and the below image is the structure of the neural network (Fig.\ref{fig:PP1}).

For the output, we use the Heatmap to roughly locate the joint position, and then use the offset to accurately adjust the joint position on the Heatmap. The heatmap (672, 28, 28) represents the feature images of 28 joints.  The offset is three times more than the heatmap feature map. It is obviously the exact location, but the three coordinates of X, Y and Z need to be located in the offset.

For the localization, the heatmap is equivalent to dividing the original image into grid points of (28,28). Each grid point represents the probability of having a joint nearby. According to the index of the Heatmap, we found the corresponding three feature graphs that accurately corrected the XYZ coordinate in offset. The value of the three feature graphs is the exact offset of the XYZ coordinate of the corresponding joint relative to the current HeatMap grid position.
\\
For example, the 3D coordinate position of the ith joint is extracted, and the corresponding feature graph is [j * 28, (j + 1) * 28 - 1], from which the position of the maximum value is found to be the grid position of the feature graph in which the joint is most likely to be. Then, query the offset based on the found Heatmap index to find the correct value. Moreover, each coordinate adds an exact offset to the position of the current grid, because xyz is separated from 24*28 feature maps on the offset, so the first index of Y and Z changes.  

As for the names of 24 joints, they are [rShldrBend,rForearmBend, rHand,rThumb2, rMid1, lShldrBend, lForearmBend, lHand, lThumb2, lMid1, lEar, lEye, rEar, rEye, Nose, rThighBend, rShin, rFoot, rToe, lThighBend, lShin, lFoot, lToe, abdomenUpper] respectively.

In addition, Since 3D keypoint data from the deep learning model will have shaking, the Low-pass filter is implemented.
\[now=prev*smooth+now*(1-smooth)\]

Basically, it combines the 6 historical frames of data for the current frame smoothing: the 0 historical frame is the current unsmoothed frame data, and then smooth forward to the last frame of the frame window, so the data after the current frame smoothing is the last frame of the window. 

\section{Result and Discussion}

\begin{table}[H]
\centering
\begin{tabular}{|ll|}
\hline
\multicolumn{1}{|l|}{GPU}                  & GTX 1070 or greater \\ \hline
\multicolumn{1}{|l|}{Unity}                & Version 2019.3.13f1 \\ \hline
\multicolumn{2}{|l|}{Barracuda}                                  \\ \hline
\multicolumn{1}{|l|}{Motion Capture Model} & Resnet34            \\ \hline
\multicolumn{2}{|l|}{3Dhuman model and city model (below)}       \\ \hline
\end{tabular}
\caption{Environment Requirement}
\label{tab:my-table}
\end{table}

Here are the environmental requirements for this project. We need to use GPU 1070 or greater, otherwise, there will be a significant latency and the motion capture performance will be very poor. Besides, we use unity as our environment and select the 2019-3-13f1 version. For the motion capture model, we selected the ResNet 34 architecture which can successfully extract the human features within the image and obtain the limb positions of the human. We also used a 3d human model and city model to get a better visualization effect. As for Barracuda, it is A lightweight and cross-platform neural network interface library for Unity that can run the neural networks on both CPU and GPU, enable users to import ML models in onnx format and fetch results. Because of this library, we can apply the ResNet 34 neural networks for our project.

\begin{figure}[]
    \centering
    \includegraphics[scale=0.7]{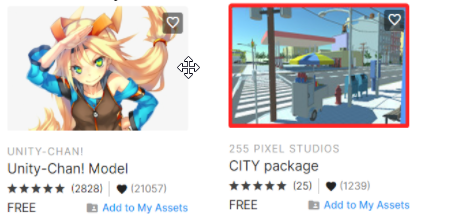}
    \caption{human model and city model}
    \label{fig:PP}
\end{figure}

\begin{figure}[]
    \centering
    \includegraphics[scale=0.7]{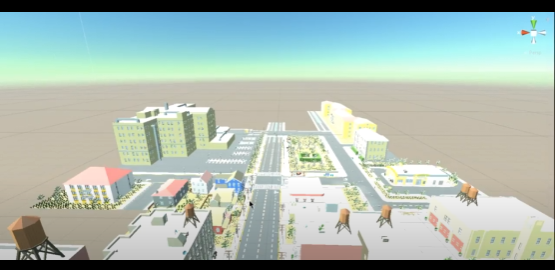}
    \caption{3D scene city model}
    \label{fig:PP}
\end{figure}

\begin{figure}[]
    \centering
    \includegraphics[scale=0.6]{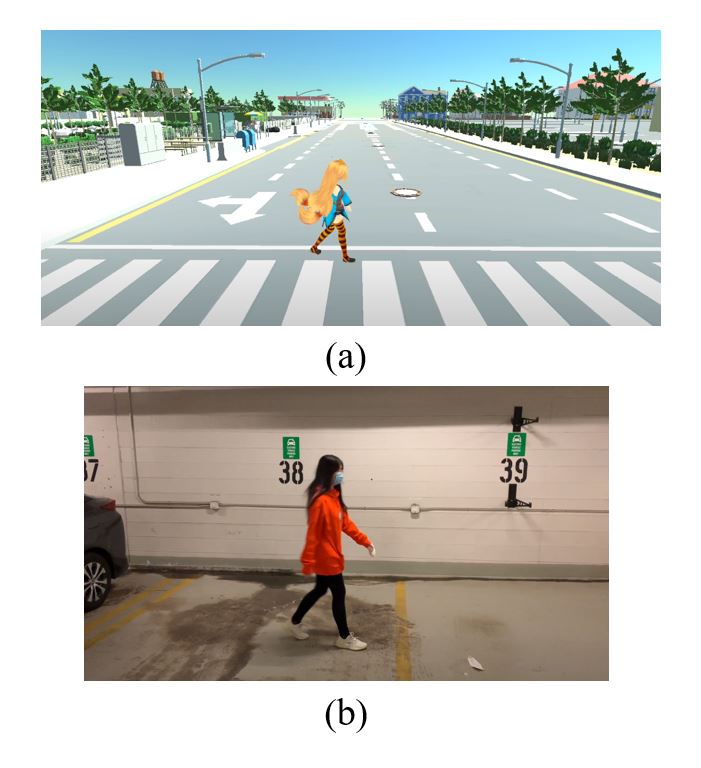}
    \caption{Real-time output result}
    \label{fig:PP}
\end{figure}

\section{Conclusion and future work}
In conclusion, we finish the implementation of real-time 3d motion capture and apply the motion data to the 3D human model and street model. In the future, we may do more research in the following areas. The first one is multi people motion capture. Because now we are focusing on a single-person motion capture on the street, but in reality, the street is way more complicated and there is more than one person, so we would like to conduct more research on multi people motion capture. Also, we would work on Improving recognition accuracy, which includes trying to capture difficult movements and trying to restore all details as much as possible.

\bibliographystyle{IEEEtran}
\bibliography{references}

\end{document}